**AI ETHICS NEEDS GOOD DATA**

Authors:

Angela Daly - Strathclyde University (Scotland)

S Kate Devitt - University of Queensland (Australia)

Monique Mann - Deakin University (Australia)

Word count: 8116

**Abstract**

In this chapter we argue that discourses on AI must transcend the language of 'ethics' and engage with power and political economy in order to constitute 'Good Data'. In particular, we must move beyond the depoliticised language of 'ethics' currently deployed (Wagner 2018) in determining whether AI is 'good' given the limitations of ethics as a frame through which AI issues can be viewed. In order to circumvent these limits, we use instead the language and conceptualisation of 'Good Data' (Daly, Devitt and Mann 2019), as a more expansive term to elucidate the values, rights and interests at stake when it comes to AI's development and deployment, as well as that of other digital technologies. Good Data considerations move beyond recurring themes of data protection/privacy and the FAT (fairness, transparency and accountability) movement to include explicit political economy critiques of power. Instead of yet more ethics principles (that tend to say the same or similar things anyway), we offer four 'pillars' on which Good Data AI can be built: community; rights; usability; and politics. Overall, we view AI's "goodness" as an explicitly political (economy) question of power (Winner 1980) and one which is always related to the degree which AI is created and used to increase the wellbeing of society and especially to increase the power of the most marginalized and disenfranchised. We offer recommendations and remedies towards implementing 'better' approaches towards AI. Our strategies enable a different (but complementary) kind of evaluation of AI as part of the broader socio-technical systems in which AI is built and deployed.

**Introduction**

Artificial Intelligence (AI) is increasingly a part of our societies and economies, principally paid for and benefiting private organisations and governments. AI applications are offered free to consumers and end-users in products such as Google Maps in exchange for access to vast amounts of data (Zuboff 2019). While AI has the potential to be used for many socially beneficial purposes, there is concern about dangerous and problematic uses of the technology, which has prompted a global conversation on the normative principles to which AI ought adhere, under the banner of 'AI ethics'. Governments, corporations and NGOs throughout the world have generated their own sets of AI ethics principles. Questions and critiques arise about the content of these ethics principles, the whether they are actually implemented, and their (legal) enforceability (Wagner 2018). Broader issues emerge about the power and privilege of the organisations, governments and individuals which are creating and implementing AI and accompanying ethical principles. For example, Google has recently announced an ethics service (Simonite 2020), yet has been mired in ethics controversies from violating privacy law (Finley 2019), working on controversial military projects (Crofts and van Rijswijk 2020) and dissolving their Ethics Board merely a week after its establishment (Statt 2019). The creation of and compliance with ethical frameworks can be expensive in time and resources, making it easier for wealthy organisations and nation-states to comply with ethical governance and even profit from it while maintaining fundamentally inequitable, unjust and self-protecting practices.

In this chapter, we argue that we need to focus more on the broader question of power and privilege than merely the aforementioned and depoliticised language of 'AI ethics'. AI ethics principles and frameworks tend to centre around the same values (fairness, accountability, transparency, privacy etc) and are insufficient to address the *justice* challenges presented by AI in society. Indeed, Amoore (2020, p. 81) contends that "a cloud ethics must be capable of asking questions and making political claims that are not already recognised on the existing terrain of rights to privacy and freedoms of association and assembly." This can be connected to arguments made by Hoffman (2019, p. 907) that a focus on a "narrow set of goods, namely rights, opportunities, and resources" is limiting in that it "cannot account for justice issues related to the design of social, economic, and physical institutions that structure decision-making power and shape normative standards of identity and behaviour." Hoffman (2019, p. 908) also contends that an "outsized focus on these goods obscures dimensions of justice not easily reconciled with a rubric of rights, opportunities and wealth" and that "emphasis on distributions of these goods fails to appropriately attend to the legitimating, discursive, or dignitary dimensions of data and information in its social and political context."

In light of these nascent critiques, we present a politically progressive approach to AI governance based on 'good data' which seeks to empower communities and progress the priorities of marginalised and disenfranchised groups worldwide. Our approach also moves the conversation beyond anthropocentrism by incorporating AI's environmental impact into normative discussions (see Foth et al. 2020). Data is the fuel for AI, providing value and power. AI capabilities are typically designed, funded, developed, deployed and regulated (if indeed at all) by the wealthy progressing the values of profit, power and dominance. AI is constructed in a way that typically reinforces and cements the status quo and existing power relationships. AI will continue to be unethical without political consciousness regarding the actors and scenarios into which it is being conceptualised, designed and implemented and the actors and scenarios that are currently excluded from consideration. Our Good Data approach instead seeks to bring these actors, issues and scenarios clearly into the spotlight and thereby into the normative conversation on AI and digital technology more generally.

Accordingly, the chapter will offer an overview and critique of AI ethics, before presenting a conceptual analysis of Good Data in the context of AI. We advance Good Data as an alternative framing for ethical (in the broad sense) questions involving digital data and conclude with some directions on Good Data an be implemented in practice vis-a-vis AI. However, for a 'Best Data' scenario for AI to be achieved, greater change contesting and replacing neoliberal capitalism may be necessary (Daly 2016; Zuboff 2019), given the political economy roots of many contemporary Bad Data practices by governments and large corporations throughout the world, which are also being implemented via AI applications. Thereby any 'quick fixes' offered by AI ethics principles may be illusory and indeed longer term, more comprehensive approach/es to 'goodness' in AI, data and society overall are needed.

**AI Ethics and Governance**

In the last few years, a global debate and discussion has emerged about governing AI and in particular whether and to which norms AI should adhere. This debate acknowledges the possibility and actuality of AI being used for normatively problematic purposes, including in physically dangerous and other harmful ways, as well as what ethical approaches humans should take towards potentially autonomous AI that may mimic our own characteristics (see e.g. Bennett and Daly 2020; Donath 2020; Dörfler 2020). A variety of stakeholders, from nation-states throughout the world to regional blocs like the European Union to large technology companies (both US- and China-based) to religious institutions have participated in this debate by issuing their own iterations of 'ethics' principles to which AI ought adhere (Daly et al. 2019). There is now also a corresponding blossoming of academic literature on AI ethics from a number of disciplines from computer science to law to philosophy to engineering examining, collating, comparing and critiquing these ethics statements and proposals (see e.g. Fjeld et al. 2020; Larsson 2020).

There have been two prominent critiques of this 'turn to ethics': one related to the form of these ethics initiatives and one relating to the substance.

One prominent critique of this 'turn to ethics' has been from Wagner (2018), who has expressed concerns about these initiatives constituting 'ethics washing' or 'ethics shopping' - that 'ethics' may 'provide an easy alternative to government regulation', in the context of strong regulation (especially containing fundamental rights protections) having encountered resistance from industry players. The majority of the ethics statements to date, even those from nation-states and other public actors, do not have legally binding force. This raises the question of how sincere and effective such principles may be, particularly in the context of a world in which large technology companies are very powerful, have engaged in problematic conduct in the past and at present, and are not always well regulated by governments (Daly 2016). It is these pre-existing 'infrastructures' and scenarios into which AI is being developed and deployed, yet these aspects are often divorced from the AI ethics discourses (Veale 2020). In addition, we see the involvement of industry players in defining these ethical principles, with the EU High-Level Expert Group a notable and controversial example as the presence of industry lobbying seemingly had an impact on the final document (see e.g. Rességuier and Rodrigues 2020). Law is one way of enforcing ethical principles but not the only one, and Hagendorff (2020) points to a broader issues with AI ethics 'usually lack[ing] mechanisms to reinforce its own normative claims', mechanisms that include law but also cover technical implementations and business process operationalisations of 'ethics'.

It is also important to note the 'weaponization' of 'ethics' in the AI debates to refer to these ethics initiatives issued mainly by governments and corporations and the 'ethics-washing' critiques which can descend into 'ethics bashing', which distracts from the more general and broad meaning of ethics as it relates to morality and virtue (Bietti 2020). In other words, 'ethics' has been used in a specific way in the AI debates both to promote (usually non-binding) lists of norms and as a target for critique that these norms are insufficient, formulated by the wrong actors and not backed up with enforcement or implementation. However 'ethics' in the more general term can make an important contribution to considerations of morality in a broad sense when it comes to AI (Bietti 2020): the ethics baby should not be thrown out with the 'AI ethics initiatives' bathwater.

The critique of ethics initiatives' substance relates to what is included and excluded as normative principles. Hagendorff (2020) identifies recurring norms in ethics declarations he analyses and compares, notably 'accountability, privacy or fairness'. Along with 'robustness or safety' Hagendorff (2020) considers these frequently occuring norms as those that 'are most easily operationalized mathematically and thus tend to be implemented in terms of technical solutions'. Hagendorff (2020) also points to the 'omissions' from many AI ethics frameworks which comprise 'red lines' on uses of AI which should be prohibited, political abuse of AI systems and the '"hidden" social and ecological costs' of AI systems. Furthermore, ethical AI principles such as fairness, accountability, transparency judge only the information systems within which AI is installed, without stepping back to analyse the socio-economic and political realities of the organisations which own and use the data - as per Bigo et al.'s (2019) *data politics* - and AI and the people who willingly or ignorantly provide the fuel to power AI.

The critique of AI ethics' substance also extends to scenarios where AI ethics are backed by legal enforceability. While a late-comer to state-led AI ethics, the US has made up for lost time since 2019 starting with the Trump Administration Executive Order on Maintaining American Leadership in Artificial Intelligence (US White House 2019). What is notable about this Executive Order is its legal enforceability albeit in the form of a direction to other US government agencies rather than e.g. a piece of general legislation containing a set of legally binding normative principles (Daly et al. 2020). The Executive Order does contain five high level principles, including the US driving the development of appropriate technical standards and protecting civil liberties and privacy (US White House 2019). However the US's policy among its own government agencies since then has promoted a deregulatory approach to AI whereby agencies should reduce regulatory barriers to AI development and adoption (Daly et al. 2020).

This approach demonstrates the limits of legal enforceability in AI ethics contexts whereby, in the US case, legal enforceability is used to mandate a deregulatory approach (Daly et al. 2020). Accordingly, "the legal enforceability of AI governance and ethics strategies does not necessarily equate to substantively better outcomes as regards actual AI governance and regulation" (Daly et al. 2020). Instead we must be alert to legal enforceability as a form of "law washing", or when the binding force of law does not in itself prevent unethical uses of AI (Daly et al. 2020).

This leads to the need to interrogate and evaluate both the form and substance of AI ethics - what they do and do not contain in substance, and what binding force they may or may not have, whether they are only 'performative' and 'instrumentalist' or whether the language of ethics is only performative. In other words, AI ethics need a Good Data approach.

**A Good Data approach**

Ethics as currently utilised in the AI debates is a limited frame through which AI issues can be viewed. While we acknowledge that ethics has a broader and more general sense than its use in AI ethics so far (Bietti 2020), we do not seek to reclaim it as a linguistic device given the term's history and tarnishment in these debates. Instead we propose 'Good Data' (Daly, Devitt and Mann 2019), as a more expansive concept to elucidate the values, rights and interests at stake when it comes to AI's development and deployment as well as that of other digital technologies. In particular, we argue that discourses, design and deployment on and of AI must engage with power and political economy, perspectives which are largely lacking in AI ethics initiatives to date (see Johnson 2019).

We conceived the notion of 'Good Data' to move beyond critique of the digital (in which we have participated and continue to do so) to the (re)imagining and articulating of a more optimistic vision of the datafied future and in particular how digital technologies and data, including but not limited to AI, can be used to further wider social, economic, cultural and political goals (Daly, Devitt and Mann 2019). We draw on work on data justice (Dencik, Hintz and Cable, 2016), data activism (Milan and van der Velden, 2016) and data politics (Bigo et al. 2019) as key elements or examples of Good Data, while our concept involves 'broader visions of goodness or ethics or politically progressive data' (Daly, Devitt and Mann, 2019).

AI ethics frameworks to date focus on evaluating the design and impacts of AI without sufficient attention on the socio-political contexts in which AI is developed and employed (as per Amoore 2020). What this means is that AI can be responsible, governable, trusted, equitable, traceable to the persons to whom it applies and reliable within the complex systems it is deployed - i.e. 'ethical' by the measure of many of the AI ethics initiatives - but the governing organisation/s responsible for the AI itself may be unethical in the broader society and environment within which it is deployed. A Good Data approach interrogates these broader situations and factors which are often absent from AI ethics initiatives.

It is reasonable to consider the scope of 'goodness' and what philosophical commitments we might have to the assertion of 'good' data. For our purposes, goodness can be a property of a thing, a service, a method, an event, a system, a process, a judgment, a sensation, a feeling or a combination of these. To identify 'good', we could suppose that there are moral facts (see Parfit 2011; Scanlon 2014). Agreement on moral facts enables standards, policies, practices and frameworks to improve information systems and communicate expectations. However, given the limits of our knowledge of moral facts (should they exist) and in light of colonial and post-colonial data practices (Arora 2016) we assume a hybrid moral theory—where we allow that some moral facts may be objective (e.g. 'tolerance' or 'openness') and others relative (e.g. Wong 1984). A hybrid theory allows respect for cultural diversity and demands case-by-case determinations of goodness and systematic values and standards. By promoting a hybrid account, we are prepared for disagreement about what is good and assume that the discovery of moral facts (if they do exist) is non-trivial and unresolved. We advocate an ethic of active seeking, openness and tolerance to diverse views on 'the good' particularly, and perhaps stridently, consultation with the underrepresented, marginalised and unheard.

**Pillars of Good Data**

Good Data contribute to understanding and justifying progressive political action by collectives. Good Data is thus situated from an ethical perspective to progress society, rather than simply satisfying an epistemic goal to inform. Therefore, we connect Good Data with political action and social justice - it means doing something *good* in the world, or equally not doing something *bad* i.e. forbearance. Good Data also can take place and be relevant to all stages in the data collection process, from the beginning to the end encompassing: when the decision is made for the data to be collected for AI use and by whom; at the point the data is collected by AI; at the point the data is processed/analysed by AI; at the point the data is used by AI; and at the point the data is reused by AI. In order to conceptualise the process and outcome of Good Data, we advance these 'pillars' on which it should rest, rather than principles to which it should apply.

We present four pillars: Community, Rights, Useability, Politics that emerge from the corpus of *Good Data* (Daly, Devitt, and Mann 2019) which can guide digital technologies and data development, including for AI. We propose that these pillars can guide an ethical and politically progressive approach to AI development, governance and implementation.

*Community*

Good Data must be orchestrated and mediated by and for individuals and collectives. Individuals and collectives should have access to, control over, and opportunities to use their own data to promote sustainable, communal living, including communal sharing for community decision-making and self-governance and self-determination (see Lovett et al. 2019; Ho and Chuang 2019). Data collection, analysis and use must be orchestrated and mediated by and for data subjects and communities advancing their data and technological sovereignty, rather than determined by those in power (Kuch et al. 2019; Mann et al., 2020). AI, constructed by communities, should be designed to assist community participation in data related decision-making and governance. This community element is usually absent from AI ethics initiatives, both in their formation - principally by elites - and their content. Commitment to theories and practices of sovereignty (Kurtulus 2004) are critical to systems architecture design including data permissions, accessibility and privacy.

### *Rights*

While we recognise that the discourse of rights is limiting (as per Amoore, 2020; Hoffman 2019), Good Data should still be collected with respect to humans and their rights, and that of the natural world, including animals and plants (Trenham and Steer 2019). The rise of big data and AI makes individual control over all their shared digital personal or community data a possibly insurmountable task. Rights language and power stems from a protection of the individual (especially in western worldviews), and there may be conflicts between the community's values and priorities with community data and the preferences of an individual within those communities. While Kalulé and Joque (2019) criticise the contemporary anthropocentrism of AI and privileging of the western human, we believe that a rights discourse is not completely futile and in principle AI can be developed to improve abidance with human rights and the rights of the environment. However such a language of rights and especially the rights of the non-human are usually absent from AI ethics initiatives. With Good Data we urge that the environmental cost and impact of AI technologies is an 'externality' which must be 'internalised' in discussions of ethics and politically progressive AI and digital data.

### *Usability*

Good Data is usable and fit for purpose, consensual, fair and transparent (Trenham and Steer 2019). Measures of fairness and other values attributed to data should be explicit (McNamara et al. 2019), and extend beyond narrowly conceived technical explanations to challenge broader structural / societal unfairness (see Hoffmann 2019 on the limits of 'fairness'). Data driven technologies must respect interpersonal relationships such as appropriate (Flintham et al. 2019), e.g. members of the same house may wish for limits on accessing each other's data - in other words, data is relational. Good Data is dependent on context, and with reasonable exceptions, should be open and published, revisable and form useful social capital where appropriate to do so (Trenham and Steer 2019). AI ethics frameworks to data dovetail with the requirements of usability, though they do fall short on the nuances of respecting interpersonal relationships and community values. There is substantial overlap between usability and community; and usability and dependence. Which is to say, that data must be usable and dependable for the community who must access and control it. If ICT systems leave communities dependent on the expertise of 'outsiders' to maintain them, then they create vulnerabilities for their sovereignty. Usability for communities is vital for all kinds of communities, from families, hospitals, schools, cultural groups, businesses and organisations as well as for nation-states).

*Politics*

Good data reveals and challenges the existing political and economic order so that data empowered citizens can secure a good polity. Citizen-led data initiatives lead to empowered citizens (see e.g. Valencia and Restrepo 2019). Open data enables citizen activism and empowerment (see Gray and Lämmerhirt 2019). Strong information security, online anonymity and encryption tools are integral to a good polity. Social activism must proceed with 'good enough data' (Gutierrez 2019; Gabrys, Prtichard and Barratt 2016) to promote the use of data by citizens to impose political pressure for social ends. How can AI contribute to the empowerment of citizens, without data, models and algorithms putting them at risk? AI systems need to be understood by citizens so that outputs and recommendations are trusted as working to their favour. To this end, the politics pillar on how activism for Good data and good AI is drawn from the other three: community, rights and usability. AI infrastructure controlled and accessed by communities that progresses their rights and interests is the gold standard of genuinely ethical AI.

Our research into Good Data encourages data optimism beyond minimal ethical checklists and duties—thus our aim is supererogatory (Heyd 1982). We recommend these Good Data pillars to progress political and social justice agendas such as citizen-led data initiatives, accepting 'good enough' data to achieve aims (Gutierrez 2019; Gabrys, Pritchard and Barratt 2016). The aim is to dismantle existing power structures through the empowerment of communities and citizens via data and digital technologies and enhancing technological sovereignty (Mann et al., 2020).

Moving away from the body of critique of pervasive 'bad data' practices by both governments and private actors in the globalised digital economy, we paint an alternative, more optimistic but still pragmatic picture of the datafied future. In order to secure a just and equitable digital economy and society we need to consider community, rights, usability and politics.

**AI for Good Data? Good Data for AI?**

But how can we implement these pillars in practice? Can AI be or be fed with or nourished by Good Data? Can AI feed and nourish Good Data? We acknowledge that here too we are not the first people to consider this issue. For instance, Floridi et al (2020) discuss the emerging 'AI for Social Good' (AI4SG) trend and formulate seven 'essential' but not 'sufficient' sociotechnical principles of their own for AI4SG. These principles are rather technocratic although Floridi et al (2020) do acknowledge the wider contexts in which AI development and deployment take place and the power imbalances which persist forming the backdrop to these developments and deployments. We instead seek to centre these contexts and power imbalances in proposing Good Data as a frame or concept for AI development.

For AI to be Good Data and Good Data to be AI, AI would need to be built for communities using data available and relevant to them. To achieve good data, communities need to gather, store and process their own data; they need to have access to open and closed data sets of relevance to their interests. Communities need cloud storage, AI classifiers, data scientists and so forth to build the tools communities need to become empowered. It is unclear what socioeconomic structures would enable genuinely ethical AI with Good Data - but certainly not current ones. At the very least it would require massive investment, democratisation and reimagining of ICT infrastructure. The most powerful produce and selectively hide data. The least powerful depend on data gathered, curated and displayed by the empowered, often data about them.

What, then, are the barriers to achieving Good Data? We have mentioned a few in passing above: the smokescreen of ethics to obscure enforceable state-led regulation; the limits of law; the broader political economy of neoliberal capitalism and corporate greed and its impacts of extractive logic on the natural world; and the corresponding power imbalances and inequalities in a world characterised by privilege and division. In addition, creating effective and trusted AI is elusive and expensive and requires access to valuable data sets, knowledge workers as well as access to high quality digital architecture, test and evaluation processes and user testing in the anticipated context of use. Data and software must be incorporated into secure and compliant back-end databases with user-friendly front end interfaces. While AI is becoming much more 'plug and play', enabling those with less skills to add data to software products and curate algorithms, the end-to-end construction of AI products to meet a societal need is still a bespoke and expensive business. The complexity of AI in addition to its cost is why AI production is dominated by three kinds of enterprise: a) technical startups, b) medium to large sized corporations and c) governments. There is comparatively very little AI constructed by low-resource community groups, non-profits, aid agencies, advocacy groups for the marginalised and disenfranchised.

What, then, can be done to address these barriers? A multipronged approach is necessary to (start to) dismantle (some of) them, with a recognition of the limits of law, code, markets and social norms (as per Lessig's (1999) modes of regulation) as tools through which Good Data can be achieved. Central though to the problem is the political economy of data in the context of neoliberal capitalism. Both governments and corporations have a strong, in some cases existential, interest in gathering, analysing and using data. Truly curbing these practices through law or ethics or markets may be extremely difficult to achieve in practice absent major societal change.

However, not to give into defeatism regarding these large challenges, we view the way forward as beginning to create an alternative vision of a datafied society and economy which promotes and achieves social and environmental justice goals, and we view incremental change for now to be the most likely pragmatic path in this direction.

There are some cases of AI for social good, for example, a software engineer developed an AI that could automatically write letters for people who received parking tickets in a way to get them a waiver from having to pay the fine (Dale 2016; DoNotPay 2020). The business aims to connect people to legal advice from parking tickets to divorce (Krause 2017). The people most likely to benefit from such an AI included those in the community who lack the funds to pay the parking ticket and may lack the education, literacy, knowledge or experience required to negotiate written legal documents. AI products for legal aid use AI to improve equity and fairness. There are cases of AI for environmental good, such as poaching deterrence and identification of rare and endangered species. AI can automatically count flocks, track animals, assess perimeter, monitor habitats.[1] Although the use of AI for surveillance -- even for ostensibly 'good' reasons -- remains controversial. Consequentialist justifications will never satisfy rights-based or duty-based obligations to other humans such as protection from persistent surveillance.

However, AI for social good is ad hoc. That is to say, private individuals generate the concept of AI to alleviate some source of injustice and proceed to develop a technology that may be useful for a specific purpose, but does not have the backing of a significant entity to ensure that AI products are chosen to improve quality of life for the most vulnerable through a process of consultation and oversight.

Non-profits and other organisations dedicated to the alleviation of human suffering and improving justice are traditionally staffed by less technical persons, such as those with legal training rather than technical training in software engineering and machine learning. It is difficult to build up the technical competence required to create AI for social justice within organisations already struggling to deliver their organisation's missions within tight budgets.

Governments might be good candidates to make AI for the good of all citizens. However, time and time again governments are found to use citizen data for uses that do not align with the values and expectations of marginalised groups within society, such as First Nations peoples (e.g. see Kukutai & Taylor 2016; Lovett et al 2019), the unemployed or marginally employed (e.g. for an overview of Australia's RoboDebt welfare surveillance program see Mann 2019; Mann 2020).

---

[1] From https://www.dronezon.com/drones-for-good/wildlife-conservation-protection-using-anti-poaching-drones-technology/

The quality of AI outputs is based on the data that it is fed and curated with. Organisations lacking access to large data sets will be unable to participate in the AI economy. Conversely, large corporations that focus on data collection as a primary asset, collect vast data sets to feed AI algorithms.

**Moving Towards 'Better' Data in AI**

We view the way forward as beginning to create an alternative vision of a datafied society and economy which promotes and achieves social and environmental justice goals, and we view incremental change for now to be the most likely pragmatic path in this direction.

As a way forward to ensuring Good Data we look to integrate Lessig's (1999) various approaches or modes of regulating technology, namely: law, code/architecture, social norms and markets with philosophical models of information, acknowledging epistemic, ethical and political conceptions of what constitutes 'the good'. In order to ensure technology such as AI rests on Good Data pillars and exhibits Good Data values, a multipronged approach involving these different modalities of regulation and conceptual apparatus is necessary.

It is insufficient to rely only on formal law to achieve ethical and politically progressive outcomes - as also recognised by Kalulé and Joque (2019), Kalulé (2019), Hoffman (2019), and Amoore (2020). We do not necessarily accept the determinist view that the law follows technological innovation which the 'regulatory disconnect' suggests (Brownsword and Goodwin 2012). Indeed, we acknowledge that digital technologies have benefitted from emerging in a period when deregulatory, neoliberal ideologies have prevailed which has led to the law playing 'catch-up' (Daly 2016).

However, we do see some potentially promising provisions such as Article 25 of the EU's General Data Protection Regulation on 'data protection by design and default', as an attempt to 'join up' law and code in implementing ethical principles. But, we and others have questions about how this translates - if indeed if it is possible to do so - into the design or hardcoding of systems (Koops and Leenes 2014), and what the consequences, including unintended of such an intention to embed principles into technological systems may be (Ihde 2006). As we also see in the US, the mere fact of AI ethics principles having the binding force of law is insufficient to establish their 'goodness'.

It is also insufficient to focus solely on social norms in the form of unenforceable ethical principles as these can indeed result in ethics-washing. We agree with Powles and Nissenbaum (2018) who see a focus on 'solving' issues of fairness, accountability and transparency as foremost among these discussions through code (and to some extent social norms) as problematic, and one which obscures the broader structural problems which are at the root of bias in machine learning. Again, Ihde's critique of the 'designer fallacy' mentioned above also applies here regarding the unintended consequences of attempting to 'design out' bias and other problems in digital technologies. Furthermore, broader existential questions about whether a particular system or technology should even be used in the first place are of key importance but also often obscured in the focus on issues such as bias (Powles and Nissenbaum, 2018). A Good Data approach to AI would certainly ask these questions before any such system was implemented, and see these problems are pertaining to broader social, political and economic contexts which will not easily be 'solved' by technology alone .

While it may not eventuate in a Good Data utopia, we view that laws, social norms, code and markets ought to promote and attempt to ensure Good Data practices. While the, or at least one, underlying problem of Good Data may be the capitalist political economy (Daly 2016; Benthall 2018), some incremental steps to promote positive change can still be taken (Raicu 2018) - or 'better' data.

Better data for AI can be promoted through a number of ways. While each alone is insufficient, together they may equate to progress. The multiplicity of ways and methods to achieve better data for AI reflect the embeddedness of AI in pre-existing, currently existing and future socio-environmental-economic conditions, and as not something that can easily be 'solved' via a statement of ethics principles.

Pragmatically, to achieve better data even the market can assist through corporate social responsibility initiatives by private sector players through ensuring they act in ethical ways (beyond legal obligations) in their product development, manufacturing, implementation and sales (Grigore, Molesworth and Watkins 2017). Better data for AI can also be advanced by environmentally sustainable corporate social responsibility initiatives (see e.g. Chuang and Huang 2018) and circular economy initiatives, and by corporations ensuring adherence to high labour standards at all stages in the supply chain. The Fairphone is an example of an attempt to produce such an ethical piece of digital technology in the private sector (Akema, Whiteman, and Kennedy 2016). Workers in technology companies can also do what they can to resist Bad Data practices, as we have witnessed at Google and Amazon in recent years (Montiero 2017; Salinas & D'Onofrio 2018). Data ethical norms and practices need to be inculcated at all levels of society including formal and informal educational settings; internet and social networking standards, media and communication channels and in the attainment of professional accreditation and qualifications. While law alone is insufficient, it also should not be dispensed with either as a tool for moving towards better data for AI.

Moreover, the debate on AI ethics has been dominated by western approaches to this topic. We also look to the Indigenous Data Sovereignty movements developed and led by First Nations peoples as presenting radically different visions of data collection and usage from the hegemonic western norm, and bring to the fore key questions of whether data should be collected and by whom (Kukutai and Taylor 2016; Lovett et al. 2019). Good Data approaches must take account of Indigenous perspectives and worldviews on data and the discrimination and oppression that Indigenous peoples and nations have hitherto experienced though western colonialism and imperialism. We are already seeing promising developments. New Zealand has recently released a draft algorithmic charter that explicitly seeks to "embed a Te Ao Māori perspective in algorithm development or procurement."[2]

**Conclusion**

We have argued here that AI needs Good Data. The four pillars of good data: community, rights, usability and politics are at the forefront of a just digital society and economy. Good Data situates genuinely ethical AI within communities and collectives, rather than individuals or large organisations. The wellbeing of the people and environment must be at the forefront of AI ethical considerations, including considerations not to use AI at all.

We have argued here that AI needs Good Data because the issues that are at the forefront of the digital society and economy go beyond pre-existing discussions of ethics. Like data itself, it is impossible for us to cover everything encompassed by 'Good Data' and accordingly we cannot offer a 'complete', 'comprehensive' or 'perfect' account of Good Data at this stage (if indeed ever).

But we can say that Good Data is a more expansive concept which aims to encompass practices beyond 'ethics' and also human rights, environmental and social justice concerns arising around data which may involve, extending beyond the focus to date on 'AI ethics' and an emerging focus on 'AI law' to address deficiencies with 'AI ethics'.

Good Data should permeate digital technology development, implementation and use at all stages in the process, and involve different tools, notably law, norms, code and markets, in order to bring about 'better' - or 'good enough' scenarios, even if the broader societal conditions and limitations mean that it is difficult to bring about 'Best Data'. Good Data can also involve the forbearance from generating and using data, either at all, or in some circumstances or by some specific people. This has implications for businesses as data collection, analysis and use should be orchestrated and mediated by, with and for data subjects, rather than determined by those in power (corporate or otherwise).

---

[2] The Draft NZ Algorithmic Charter is available here: https://data.govt.nz/use-data/data-ethics/government-algorithm-transparency-and-accountability/algorithm-charter/

We also hope that Good Data can encompass a more global approach, rather than just (re)centring perspectives from the Global North, as already noted – and critiqued - by Arora (2016) and Kalule and Joque (2019). However, we acknowledge we are also coming from a northern/western perspective ourselves.[3] Already there is emerging discussion from China in particular on technology ethics, and legislative activity in many jurisdictions around the world regarding data localisation (Melashchenko 2019). The Indigenous Data Sovereignty movements also display different worldviews and approaches to issues of data situated in Indigenous laws, cultures and traditions, countervailing the practices and uses of data by colonial and imperial forces against Indigenous peoples, and representing more perspectives beyond the western focus on normativity and ethics as regards technologies including AI.

Pragmatically, we view the next steps for all involved in the digital society and economy (which, in fact, is all of us) as trying to engage and empower each other to build Good Data initiatives and communities of change, rather than letting governments and corporations build a Bad Data future for us. Yet it is also important that governments and corporations contribute positively to the Good Data future by taking note and implementing 'good' and more ethical data practices. Only with such a multifaceted approach encompassing will we be able to achieve some semblance of Good Data for AI and for the digital more generally.

---

[3] Or perhaps more accurately for two of us, a 'Global North-in-South' perspective - see Mann and Daly (2018).

Daly, A., Devitt, S.K. & Mann, M. (2021). AI Ethics Needs Good Data. Pieter Verdegem (ed), *AI for Everyone? Critical Perspectives*. University of Westminster Press. [under peer review]

Daly, A., Devitt, S.K. & Mann, M. (2021). AI Ethics Needs Good Data. Pieter Verdegem (ed), *AI for Everyone? Critical Perspectives*. University of Westminster Press. [under peer review]

Trenham, C. and Steer, A. 2019. The Good Data Manifesto. In A. Daly, S.K. Devitt and M. Mann (Eds.), *Good Data*, pp. 37-53. Amsterdam: Institute of Network Cultures.

US White House. 2019. Executive Order on Maintaining American Leadership in Artificial Intelligence. 11 February. Retrieved from: https://www.whitehouse.gov/presidential-actions/executive-order-maintaining-american-leadership-artificial-intelligence/

Veale, M. 2020. A Critical Take on the Policy Recommendations of the EU High-Level Expert Group on Artificial Intelligence. *European Journal of Risk Regulation,* 1-10. DOI: 10.1017/err.2019.65

Wagner, B. 2018. Ethics as an escape from regulation: From ethics-washing to ethics-shopping. In: E. Bayramoglu, I. Baraliuc, L. Janssens et al. (Eds.), *Being Profiled: Cogitas Ergo Sum: 10 Years of Profiling the European Citizen*, pp. 84-98. Amsterdam: Amsterdam University Press.

Winner, L. 1980. Do Artefacts Have Politics? *Daedalus*, 109(1), 121-136.

Wong, D.B. 1984. *Moral Relativity*. Berkeley, CA: University of California Press

Zuboff, S. 2019. *The Age of Surveillance Capitalism: The Fight for a Human Future at the New Frontier of Power*. London: Profile Books.
20